\documentclass[letterpaper]{article} % DO NOT CHANGE THIS
\usepackage{aaai24}  % DO NOT CHANGE THIS
\usepackage{times}  % DO NOT CHANGE THIS
\usepackage{helvet}  % DO NOT CHANGE THIS
\usepackage{courier}  % DO NOT CHANGE THIS
\usepackage[hyphens]{url}  % DO NOT CHANGE THIS
\usepackage{graphicx} % DO NOT CHANGE THIS
\usepackage{natbib}  % DO NOT CHANGE THIS AND DO NOT ADD ANY OPTIONS TO IT
\usepackage{caption} % DO NOT CHANGE THIS AND DO NOT ADD ANY OPTIONS TO IT
\usepackage{algorithm}
\usepackage{algorithmic}
\usepackage{amsmath}
\usepackage{multirow}
\usepackage{booktabs}       % professional-quality tables
\usepackage{graphicx}
\usepackage{amssymb}
\usepackage{enumitem}
\usepackage{pifont}
\usepackage{xcolor}
\usepackage{tablefootnote}
\usepackage{newfloat}
\usepackage{listings}
\DeclareCaptionStyle{ruled}{labelfont=normalfont,labelsep=colon,strut=off} % DO NOT CHANGE THIS
\floatstyle{ruled}
\newfloat{listing}{tb}{lst}{}
\floatname{listing}{Listing}
\title{Tackling Vision Language Tasks Through Learning Inner Monologues}\author {
    Diji Yang\textsuperscript{\rm 1},
    Kezhen Chen\textsuperscript{\rm 2},
    Jinmeng Rao\textsuperscript{\rm 2}, \\
    Xiaoyuan Guo\textsuperscript{\rm 2},
    Yawen Zhang\textsuperscript{\rm 2}, 
    Jie Yang\textsuperscript{\rm 2}, 
    Yi Zhang\textsuperscript{\rm 1}
}
\affiliations {
    \textsuperscript{\rm 1}University of California, Santa Cruz, 
    \textsuperscript{\rm 2}Mineral\\
    \{dyang39,yiz\}@ucsc.edu \\ 
    \{kezhenchen, jinmengrao, xiaoyuanguo, yawenz, yangjie\}@mineral.ai
}
\usepackage{bibentry}
\begin{document}

\maketitle

%%%%%%%%%%%%%%%%%%%%%%%%%%%%%%%%%%%%%%%%%%%%%%%%%%%%%%%%%%%%%%
\begin{abstract}

Visual language tasks such as Visual Question Answering (VQA) or Visual Entailment (VE) require AI models to comprehend and reason with both visual and textual content. Driven by the power of Large Language Models (LLMs), two prominent methods have emerged: (1) the hybrid integration between LLMs and Vision-Language Models (VLMs), where visual inputs are firstly converted into language descriptions by VLMs, serving as inputs for LLMs to generate final answer(s); (2) visual feature alignment in language space, where visual inputs are encoded as embeddings and projected to LLMs' language space via further supervised fine-tuning. The first approach provides light training costs and interpretability but is hard to be optimized in an end-to-end fashion. The second approach presents decent performance, but feature alignment usually requires large amounts of training data and lacks interpretability. 

To tackle this dilemma, we propose a novel approach, \textbf{I}nner \textbf{M}onologue \textbf{M}ulti-Modal \textbf{O}ptimization (IMMO), to solve complex vision language problems by simulating inner monologue processes, a cognitive process in which an individual engages in silent verbal communication with themselves. More specifically, we enable LLMs and VLMs to interact through natural language conversation (i.e., inner monologue) and propose to use a two-stage training process to learn how to do inner monologue (self-asking questions and answering questions). IMMO is evaluated on two popular tasks and achieves competitive results compared with hybrid integration approaches, while it uses significantly less training data and provides greater interpretability compared with embedding alignment approaches. The results suggest by emulating the cognitive phenomenon of internal dialogue, our approach can enhance reasoning and explanation abilities, contributing to the more effective fusion of vision and language models. More importantly, instead of using predefined human-crafted monologues, IMMO learns this process within the deep learning models, promising wider applicability to many different AI problems beyond vision language tasks.
%we can view chain-of-thoughts as a form of inner monologue with predefined human-crafted monologues, while our approach learns this process within the deep learning models, promising wider applicability to many different AI problems beyond vision language tasks. 

\end{abstract}
%%%%%%%%%%%%%%%%%%%%%%%%%%%%%%%%%%%%%%%%%%%%%%%%%%%%%%%%%%%%%%

\section{Introduction}
\begin{figure}[tbh]
    \centering
    \includegraphics[width=0.85\linewidth]{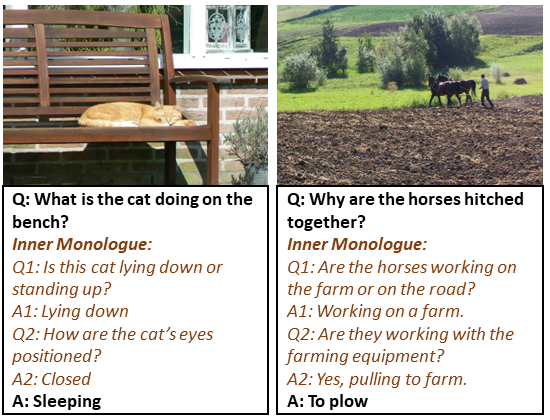}
    % \vspace{-2ex}
    \caption{Examples of multi-model inner monologue.
    }\label{fig:example}
\end{figure}

Evidence shows that explicitly using natural language as the intermediate representation of reasoning is effective and essential for human cognition \citep{Gentner2003, Forbus2017, Crouse2018, tpn2f, rscan, zhang2023multimodal}. Recently, large language models (LLMs) have achieved substantial advancements. Notable models like PaLM~\citep{palm}, InstructGPT~\citep{ouyang2022training}, and LLaMA~\citep{touvron2023llama} showcase their immense potential in the field of natural language processing and commonsense reasoning. Many researchers explore using natural language as the intermediate representation to bridge multiple modalities. For instance, various modalities, including the vision modality, can be first projected to the natural language space, and then LLMs are utilized to perform multi-modality understanding via language processing. Two research directions to add visual inputs into language space have been actively studied recently. The first direction is the hybrid integration between vision-language models (VLMs) and LLMs \citep{yang2022empirical, salaberria2023image, chatcaptioner}. Hybrid integration approaches aim to enable LLMs to utilize VLMs in a zero-shot or few-shot manner, i.e., LLMs act as a central reasoner or planner and VLMs serve as tools or sensors. These models do not require heavy training costs and provide interpretability as the model outputs from LLMs and VLMs are transparent. However, as LLMs do not access the visual inputs directly, they may miss some visual details in the images. Also, most hybrid integration approaches merge LLMs and VLMs in a discrete space, which are hard to be optimized in an end-to-end fashion. The second direction is embedding alignment \citep{dai2023instructblip, chen2023x,li2023otter, liu2023visual}. Visual information is transformed into visual embeddings, which are then mapped onto the language embedding space and employed as input embeddings for LLMs. Then, the LLMs are tuned via supervised fine-tuning to fuse vision and language information together to achieve decent performance. However, learning the cross-modality embedding alignment heavily rely on the training data. To get decent performance, users need to prepare a large number of high-quality data containing accurate domain knowledge and complex reasoning, which requires heavy engineering efforts and high curation costs. Furthermore, during the inference time, the entire model remains a black box, making interpretation difficult \citep{9879942}. The heavy costs associated with collecting cross-modality data, combined with low interpretability, limit the applicability of this approach in many domains in practice. 

To address the challenges, we introduce a novel approach, \textbf{I}nner \textbf{M}onologue \textbf{M}ulti-Modal \textbf{O}ptimization (IMMO), to solve complex vision language problems by simulating the inner monologue process, a cognitive process in which an individual engages in silent verbal communication with themselves. When people solve complicated reasoning problems, they tend to use ``inner monologue" by performing reasoning via multi-turn self-conversations in their minds. Inner monologue helps people organize their thoughts and work through the optimal answers as a form of problem-solving \citep{inner_monologue, inner_monologue_llm}. Inspired by this process, we enable LLMs and VLMs to interact through natural language conversation and propose to use supervised learning and reinforcement learning to learn how to perform the inner monologue. We choose one LLM as the \textit{Reasoner} and one VLM as the \textit{Observer}. Given a visual input and a visual reasoning problem, the Observer perceives the visual information and abstracts it to a natural-language description. The Reasoner decides whether the description has sufficient information to solve the problem or generates an advanced question for the Observer to acquire more visual information. With multiple turns, the Reasoner organizes the information and works through an answer. Figure~\ref{fig:example} shows two examples of solving visual questions with the inner monologue. To automatically learn the process of the inner monologue, the entire system is optimized by a policy gradient method named Proximal Policy Optimization (PPO) \citep{schulman2017proximal}. 
%Since the system-level optimization involves updating two models, we use an optimization scheme that alternatively trains the Reasoner and the Observer. In each epoch, we activate either the Reasoner or the Observer as the policy network, while the other model is used as the environment model to provide feedback. Only the parameters of the activated model are optimized at each epoch and the environment model is kept frozen.

In summary, our contributions are as follows:
\begin{itemize}
\item Inspired by human cognition, we propose a novel approach IMMO for vision-and-language reasoning. IMMO includes a Reasoner model and one or more Observer model(s) to communicate with each other in natural language, significantly improving their ability to solve complex problems together. IMMO can be trained efficiently and has interpretability. This inner monologue-based approach is flexible and can be adapted to other modalities or models. 
\item We propose a two-stage training process to let observer(s) and Reasoner learn how to work together: first using annotated multi-turn reasoning data to train the initial models, then adopting reinforcement learning to further improve the models. We created a new human-like multi-turn inner monologue reasoning training corpus by augmenting existing VQA data with GPT-3.5
%\footnote{The codes and data will be published upon acceptance.}
. We find this low-cost training corpus effective.

\item We evaluate IMMO on two vision-language reasoning tasks. Experiments show that IMMO can achieve competitive results compared with hybrid integration approaches, while it uses significantly less training data and provides greater interpretability compared with embedding alignment approaches.

% Experiments show that our approach can achieve competitive results compared with state-of-the-art models. Compared with alignment approaches, IMMO used significantly fewer training data and strong interpretability. Compared with hybrid approaches, it achieved better performance. 

\end{itemize}

\begin{figure*}[tbh]
    \centering
    \includegraphics[width=0.85\linewidth]{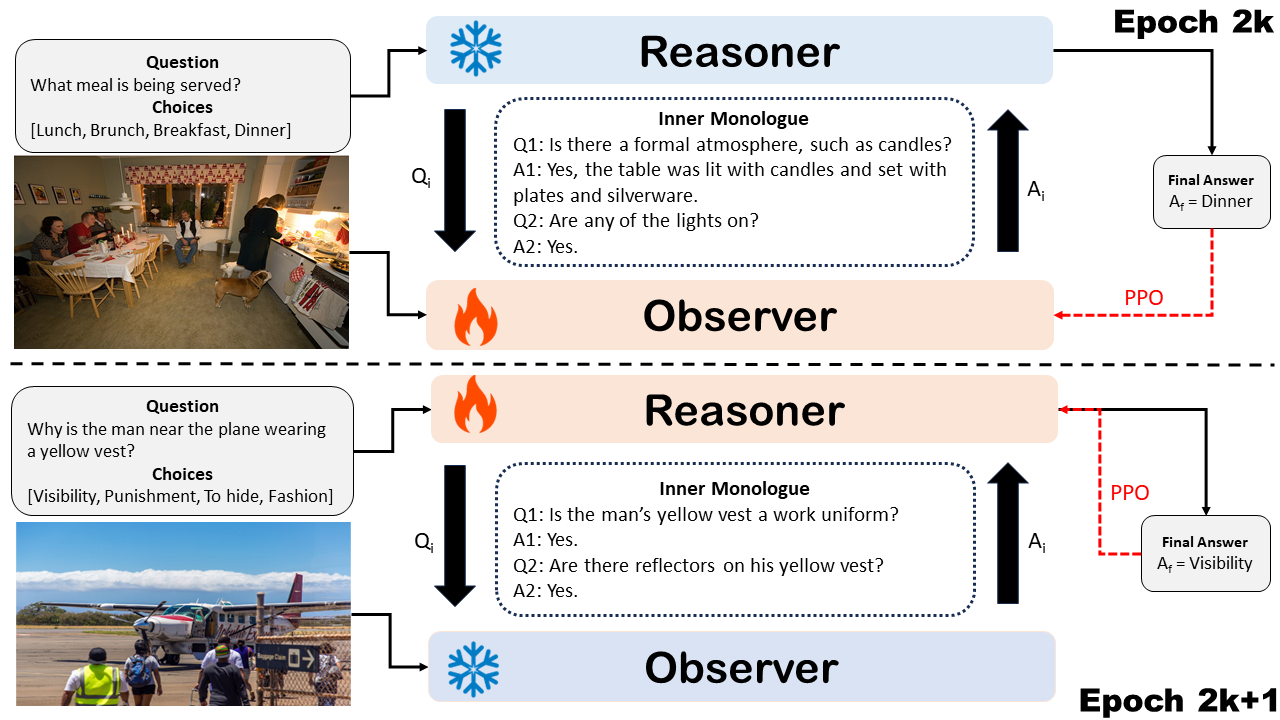}
    % \vspace{-2ex}
    \caption{The IMMO framework automatically acquires inner monologue capabilities through reinforcement learning. In each training epoch, the Reasoner (LLM) and Observer (VLM) are alternately designated as the actively trainable model, highlighted with a pink hue and a fire icon, while the other model assumes the role of a static environmental representation, distinguished by a light blue shade and a snow icon. During the ($2k$)-th epoch, the Reasoner functions as the fixed environmental model, while the Observer becomes the focus of updates. Subsequently, in the ($2k+1$)-th epoch, the roles shift, with the Reasoner now undergoing updates as the active model, and the Observer assuming the position of the static environmental representation. PPO policy gradients are used for iterative updates of the model parameters. }
    \label{fig:rl}
\end{figure*}

%%%%%%%%%%%%%%%%%%%%%%%%%%%%%%%%%%%%%%%%%%%%%%%%%%%%%%%%%%%%%%
\section{Related Works}

The success of large pre-trained language models (LLMs) has led to significant advancements in solving vision-language problems by fusing visual representations into the language space. The essence of these works is to allow LLMs to understand information from other modalities, using their rich pre-trained language knowledge and the emerging ability \citep{wei2022emergent}. Several recent works have explored two research directions, embedding alignment and hybrid integration. Approaches with embedding alignment focus on projecting visual embeddings to the language space and fusing vision and language information via supervised fine-tuning in the language space \citep{dai2023instructblip, chen2023x,li2023otter, liu2023visual}. These models learn the embedding projection from perception models to LLMs with a large amount of high-quality vision-language instruction-tuning data.
%For example, LLaVA model \citep{liu2023visual} learns an embedding projection layer to map the visual embeddings from a vision transformer to language input embeddings. Then, an LLM is fine-tuned using machine-generated dialogues to enable LLMs to follow instructions. Otter \citep{li2023otter} adapts the OpenFlamingo \citep{openflamingo} architecture and maps visual embeddings to the language space at each transformer block. Otter is fine-tuned on a large-scale image-text dataset to enhance visual alignment and reasoning. 
Despite the impressive performance, these models demand extensive engineering efforts to collect the training data and struggle with interpretability, given the difficulty for human to understand latent embeddings and comprehend the reasoning process of deep models.
%Users may find it difficult to determine whether failure cases arise from a lack of knowledge of the language model or misrepresenting visual information.
Our approach focuses on converting visual inputs to language descriptions while keeping decent performance, which provides more interpretability and reduces training costs significantly. Approaches involving hybrid integration convert visual inputs to language descriptions, such as image captions, via VLMs and solve problems with LLMs~\citep{yang2022empirical,salaberria2023image}. However, this approach may lead to captions that are irrelevant to the question. To address this problem, several works adapt interactive multi-turn conversations to promote VLMs and LLMs interacting with each other and acquiring more information ~\citep{idealgpt,chatcaptioner}. Despite these works providing more interpretability and accessibility, they are usually in zero-shot or few-shot settings and have significantly lower performance compared to the embedding-alignment-based approaches. Our approach introduces a novel framework to optimize hybrid integration systems, which gives a more decent performance while preserving interpretability.

Another branch of research related to our work is multi-agent learning. Researchers in multi-agent collaborations have studied communication and dialog between models \cite{foerster2016learning}. Such interactions enable models to exchange information, clarify doubts, negotiate strategies, and collaborate towards a common goal. They can share experiences, strategies, and insights gained from different perspectives. Multi-Agent Reinforcement learning has been studied in various applications such as the games of Go, robotics, and autonomous driving. Depending on whether the agents are fully collaborative, fully competitive, or a mix of the two, one of the two approaches, Markov/stochastic games, and extensive-form games, are usually used \cite{zhang2021multi}. The RL method we proposed is motivated by these earlier works, while our study focuses on agents that can output natural language and facilitates model-to-model communication using \textbf{natural language} with the help of the recent advance in large-scale language models.

%%%%%%%%%%%%%%%%%%%%%%%%%%%%%%%%%%%%%%%%%%%%%%%%%%%%%%%%%%%%%%
\section{Approach}

An overview of the IMMO framework for solving complex vision and language problems is shown in Figure~\ref{fig:rl}. Our framework contains two components: Reasoner and Observer. The ``inner monologue" process in the IMMO is described as follows. 

The Observer takes images as the inputs and generates textual descriptions to describe the key information it observes. The Reasoner takes the generated textual descriptions and performs reasoning by either generating a new query for the Observer or generating the final results of the task. We choose an LLM as our Reasoner model and a VLM as our Observer model. The objective of the Reasoner is to generate effective queries to obtain targeted information and the objective of the Observer is to provide correct information based on the queries from the Reasoner. With multi-turn querying-answering conversations between the Reasoner and the Observer, the Reasoner gathers information to address the vision and language problems. Meanwhile, the Observer receives the queries and perceives more visual details from the image. 

In this section, we start by presenting the IMMO framework and then introduce the two-stage training approach for the IMMO framework.

% To achieve these goals, our methodology consists of two steps:
% \begin{itemize}
%   \item \textbf{Step 1: Human-prior fine-tuning}: We utilized a high-quality training corpus containing multi-turn visual question-answering conversations to warm up the Reasoner and the Observer. With this fine-tuning stage, both models can rapidly capture the human-like reasoning process (Section~\ref{sec:sl}).
%   \item \textbf{Step 2: Inner Monologue Optimization}: After the previous stage, the whole system is optimized using result-driven reinforcement learning. We employ alternate training to balance the capabilities of both models (Section~\ref{sec:rl}).
% \end{itemize}

% Each step is discussed in the following sections.

\subsection{Inner Monologue Multi-Modal Optimization}

In the IMMO framework, the Reasoner and the Observer work together to solve a problem. Initially, the Observer receives the Image $I$ and generates a caption $C$ to describe the basic visual information in the image. A text container $IM$ will be used here to track the inner monologue, including the initial caption $C$:

\begin{equation}
IM_0 = C = Observer(I)
\label{eq:state0}
\end{equation}

At the intermediate $i$-th turn ($i \in [1, t]$ where t is the predefined maximum conversation turn), the Reasoner and the Observer will interact. Firstly, the Reasoner receives the original textual description of the problem/task $P$ combined with $IM_{i-1}$ and generates a query $Q_i$. Then, given $Q_i$, the Observer will provide the answer $A_i$ based on the image. This process is shown as the equations~\ref{eq:state1}, and~\ref{eq:state2}.

\begin{equation}
Q_i = Reasoner(P, IM_{i-1})
\label{eq:state1}
\end{equation}

\begin{equation}
A_i = Observer(I, Q_i)
\label{eq:state2}
\end{equation}

At the end of each turn, both question and answer that are generated within the current turn will be added to the inner monologue history. The $IM_i$ is defined as:

\begin{equation}
IM_i = C + \sum_{j=0}^{i} (Q_j + A_j)
\label{eq:state3}
\end{equation}

After the final conversation turn $t$, the Reasoner will provide its prediction $A_{f}$ for the original problem $P$ based on all the collected inner monologue: 

\begin{equation}
A_f = Reasonser(P + IM_t)
\label{eq:state4}
\end{equation}

% \begin{equation}
% Q_1 = Reasoner(Concat(P, C))
% \label{eq:state1}
% \end{equation}

% \begin{equation}
% A_i = Observer(I, Q_i)
% \label{eq:state2}
% \end{equation}

% \begin{equation}
% Q_{i+1} = Reasoner(Concat(P, C, A_1, ..., A_i))
% \label{eq:state3}
% \end{equation}

% \begin{equation}
% A_t = Observer(I, Q_t)
% \label{eq:state4}
% \end{equation}

% \begin{equation}
% A_{f} = Reasoner(Concat(P, C, A_1, ..., A_t))
% \label{eq:state5}
% \end{equation}

During the multi-turn iteration, the Observer only accesses the input image and the most recent information query generated by the Reasoner, while the Reasoner can access the complete QA history at any given timestamp through the input prompt. Once the interaction reaches a pre-defined number of turns, the system prompts LLM for the final prediction. Next, we describe how IMMO optimizes the Reasoner and the Observer jointly.

\subsection{Two-Stage Training}
\label{Sect2Stage}
Although this multi-turn conversational framework can be used in a zero-shot manner through prompting, the collaboration between the Reasoner and the Observer is suboptimal. The collaboration between the Reasoner and Observer should be further improved. For example, the Reasoner needs to be familiar with the Observer's capability in order to generate appropriate queries that the Observer can answer. The Observer, on the other hand, should be optimized to extract correct visual information based on the queries from the Reasoner.

To alleviate the aforementioned underperformed collaboration problem, IMMO uses a two-stage training process. First, high-quality inner monologue conversational data is used for supervised fine-tuning (SL) of both the Reasoner and the Observer. Second, reinforcement learning (RL) is used for further optimization. Figure~\ref{fig:rl} shows the overview of the IMMO framework, and only the system-level reinforcement learning is illustrated. 

\subsubsection{Supervised Human-prior Fine-tuning} \label{sec:sl}

To provide the Reasoner and the Observer a better starting point for reinforcement learning in the next stage, we employ supervised fine-tuning, similar to the approach used in InstructGPT \citep{ouyang2022training, cruz2017pre}. 
Our training process focuses on imparting effective inner monologue to the model, going beyond simple chit-chat or prompt-based zero-shot learning. To achieve this, we enhance our pre-trained language model by introducing human prior knowledge and reasoning patterns with supervised fine-tuning. We utilize high-quality multi-turn conversational question-answering pairs annotated by humans as our training data. Both the Reasoner and the Observer are trained on the human-annotated data as a warm-up process.

In order to impart human reasoning patterns to the language model, we employ the instruction fine-tuning method \citep{chung2022scaling} in this training step. Inspired by Chains-of-Thought prompting \citep{wei2022chain} in the reasoning task, this stage trains the model to mimic the correct thinking path. Instead of directly answering the basic question, the correct pattern involves reasoning through multiple turns of QA pairs and then arriving at the final answer. 

\begin{algorithm}[tbh]
\textbf{Dataset}: (Problem $P$, Image $I$, Ground Truth $G$) tuples \\
\textbf{Reasoner}: a pre-trained large language model \\
\textbf{Observer}: a pre-trained vision-language model \\
%\textbf{Concat}: string concatenation \\
\textbf{N}: training epoch \\
\textbf{t}: pre-defined max turns\\
\textbf{k}: any none-negative integer
\caption{IMMO Reinforcement Learning}
\label{alg:RL}
\begin{algorithmic}[1]
\FOR{epoch = 1 to N}
    \STATE Set $Reasoner$ as the active model $\mathcal{M}$
    \STATE Set $Observer$ as the environment model $\mathcal{E}$
    \IF {epoch = 2k}
        \STATE Set $Observer$ as the active model $\mathcal{M}$
        \STATE Set $Reasoner$ as the environment model $\mathcal{E}$
    \ENDIF
    \STATE Sample ($P$, $I$, $G$) from the dataset
    \STATE $C \gets Observer$($I$)
    \STATE Set $IM_0 = C$
    % \STATE Set $i = 1$
    % \STATE $Q_1 \gets Reasoner$($Concat(P, C)$)
    % \STATE Set $i \gets i + 1$
    \FOR{$i=1$ to $t$}
        \STATE $Q_{i} \gets Reasoner(P, IM_{i-1})$
        \STATE $A_i \gets Observer(I, Q_i)$
        \STATE $IM_i = IM_{i-1}+Q_i+A_i$
    \ENDFOR
    \STATE $A_{f} = Reasoner(P, IM_t)$
    \STATE Reward $\gets \mathcal{R}$ \COMMENT{Eq.~\ref{eq:final-reward}}
    \STATE Update $\mathcal{M}$ using PPO
\ENDFOR
\end{algorithmic}
\end{algorithm}

\subsubsection{Reinforcement Learning} \label{sec:rl}

IMMO uses a special alternative training process for system-level reinforcement learning to jointly optimize multiple models while taking into account the dynamic interactions between models. Since the system involves two models, we use the alternating training strategy to prevent issues that may arise from updating two models simultaneously, such as the imbalance of capabilities of the Reasoner and the Observer~\citep{goodfellow2014generative}. Specifically, at the $2k$-th epoch where $k$ is a non-negative integer, we set the Observer as the active model (policy network) and the Reasoner as the environment model to provide feedback; at the $2k+1$-th epoch, we switch the Reasoner to be active and change the Observer as the environment model. 
% Specifically, at the $ (2k+1) $-th epoch where $k$ is a non-negative integer, we set the Reasoner as the active model (policy network), and the Observer as the environment model to provide feedback. At the $2k$-th epoch, the active model is switched to the Observer and the Reasoner becomes the environment model. 
During training, we only update the active model. Following the common approaches used in previous works \citep{stiennon2020learning, ziegler2019fine, ouyang2022training, vonwerra2022trl} for fine-tuning auto-regressive decoder-only generative model, we treat the active model as the policy networks. The active model is updated by PPO~\citep{schulman2017proximal} and the environment model remains frozen. Notably, the active model and the environment model only affect which model will be updated, and the input/output of each model strictly follows the multi-turn framework as shown in Figure \ref{fig:rl}. The algorithm uses the exact matching between the predicted answer $A_{f}$ and the ground-truth answer $G$ as the major reward factor: % as shown in the equation~\ref{eq:major-reward}. 
\begin{equation}
r(A_{f}, G) =
\begin{cases}
& 1 \quad \textrm{if} \quad A_{f} = G\\
& 0 \quad \textrm{otherwise}
\end{cases}
\label{eq:major-reward}
\end{equation}

The final reward $R$ shown in equation~\ref{eq:final-reward} also includes a KL penalty~\citep{jaques2017sequence} weighted by $\beta$ to ensure that the updated model $\mathcal{M}$ does not deviate too far from the well-trained starting point $\mathcal{M}_0$~\citep{ziegler2019fine}. 

\begin{equation}
    R = r(A_{f}, G) + \beta KL(\mathcal{M}, \mathcal{M}_0)
\label{eq:final-reward}
\end{equation}

The training goal is to optimize the policy that maximizes the expected reward. The overall training procedure is shown in Algorithm~\ref{alg:RL}.

%%%%%%%%%%%%%%%%%%%%%%%%%%%%%%%%%%%%%%%%%%%%%%%%%%%%%%%%%%%%%%
\section{Experiment}
To evaluate the effectiveness of IMMO for complex vision-language reasoning, we conducted experiments on two popular tasks: Commonsense Visual Question Answering (VQA) and Visual Entailment (VE). Both tasks require models to have commonsense knowledge and reasoning abilities. This section first describes our implementation of the IMMO framework, then presents the details of these two tasks.

\subsection{Data and Implementation}
\begin{figure}[h]
    \centering
    \includegraphics[width=\linewidth]{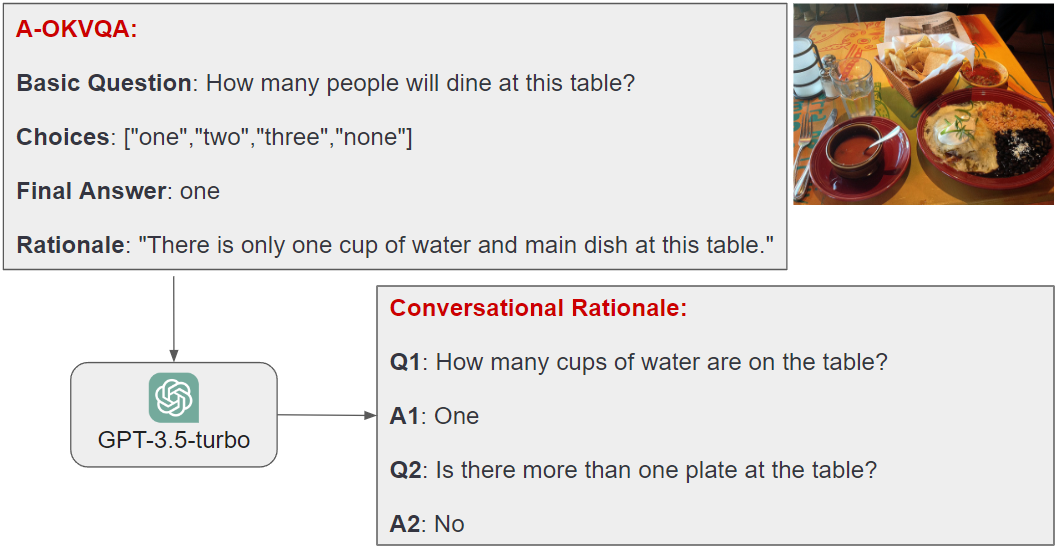}
    % \vspace{-2ex}
    \caption{Example of converting human written declarative rationale to dialogue form reasoning path.
    }\label{fig:rationale}
\end{figure}

% \subsubsection{Training}
We construct a new training corpus for supervised human-prior fine-tuning by utilizing the A-OKVQA~\citep{schwenk2022okvqa} dataset, which includes human-annotated reasoning paths labeled as rationale. We derive inner monologue from rationale. As demostrate in the Figure~\ref{fig:rationale}, by prompting GPT-3.5 in a zero-shot manner, we transform rationale into two-turn question-answering pairs. The results are then combined with 17k single-turn VQA samples.
%leveraging the LLM's natural language ability. Based on \citet{liu2023visual}'s prompt, we prompted GPT-3.5 \citep{ouyang2022training} in a zero-shot manner to augment 17k multi-turn visual question-answering samples.
Each sample in the training corpus contains a question, a choice list, two rounds of QA conversations, and the correct answer. At the supervised fine-tuning stage, we optimize the autoregressive LLM by performing the next token prediction task over this augmented corpus. At the reinforcement learning stage, the training is mainly based on the Transformers-Reinforcement-Learning (TRL) solution ~\citep{vonwerra2022trl} to wrap up the Hugging Face trainer ~\citep{wolf-etal-2020-transformers}. For different tasks (VQA or VE), reinforcement learning is performed on task-specific training sets.
% At each epoch, one model serves as the trainable active policy network, and the other acts as a frozen environment model for inference only. This configuration last for only one training epoch, and will be swapped in the next epoch. 

% \subsubsection{Models}
Our proposed system uses the Vicuna-7b~\citep{vicuna2023} language model and BLIP-2~\citep{li2023blip} vision-language model. To ensure computational efficiency, we employed the Low-rank adaptation (Lora)~\citep{hu2021lora} to train only 0.06\% of the Vicuna-7b model, which corresponds to 5 million parameters. Our experiments primarily focus on the validation of the methodology. For broader applicability, we chose a model that can be trained on a single NVIDIA A100-40G GPU or equivalent, instead of a more powerful but larger model. Task-specific prompts for both LLM and VLM were designed manually, inspired by prompt templates used by~\citet{idealgpt,liu2023visual}.

\subsection{Commonsense Visual Question Answering}

We conduct our experiments on the ScienceQA (SQA) ~\citep{lu2022learn} dataset, which is a standard benchmark for commonsense visual question answering. It consists of 21k QA pairs collected from elementary and high school courses, with 48.7\% of the questions including images, making it a VQA task. Success on this dataset requires appropriate commonsense knowledge as well as reasoning skills. We follow the official train/validation/test split for all our experiments.

\paragraph{Baselines}
To study the impact of RL optimization and inner monologue on model performance, we conduct experiments with 3 baselines. As demonstrated in Table~\ref{tab:result-sqa}, different training methods and reasoning aids were examined, while we use Vicuna-7B as LLM and BLIP-2 as captioning model in all cases. The first baseline is PICa~\citep{yang2022empirical}, which first proposed LLM plus image captions for Knowledge Base VQA. For a fair comparison, instead of using the default GPT-3 as LLM in PICa, we let PICa use Vicuna as its LLM. The second baseline is Vicuna-16-shots, which incorporates the Chain-of-Thoughts (CoT) prompting~\citep{wei2022chain} with the Vicuna model. The third baseline is Vicuna-SL, which is LLM fined tuned with the training data following instruction fine-tuning~\citep{chung2022scaling, alpaca}. We also compare two training methods for IMMO: few-shot learning (IMMO 16-shots) vs. two-stage training described in Section \ref{Sect2Stage} (IMMO SL+RL). 

\subsubsection{Results}
Following the baseline setting, table~\ref{tab:result-sqa} presents our results on the SQA test set. PICa, as a zero-shot baseline, achieves a modest accuracy of 54.3\%. By using 16 in-context examples and CoT, Vicuna-16-shots attain an accuracy of 68.6\%. With the same LLM model and in-context examples, enabling inner monologue with only few-shot learning (IMMO-16-shots) improves the performance by 5.4\% (from 68.6\% to 74\%). With the two-stage training,
%supervised fine-tuned human-reasoning corpus, then reinforcement learning on SQA training set, 
IMMO further improves and achieves 84.8\% accuracy, which is 6.5\% over Vicuna-SL. Our experiment results highlight the role of inner monologue in facilitating reasoning, while also demonstrating how RL training can enhance the overall system's capabilities.

\begin{table}[tb]
\centering
\resizebox{1\linewidth}{!}{
  % \centering
  \setlength{\tabcolsep}{3pt}
  \begin{tabular}{l l l c }
    \toprule
     Method~\tablefootnote{All the experiments in the table are named after the methods, and the language models involved are all Vicuna-7B.}
     & Training & Reasoning Aids & Accuracy(\%) \\
    \midrule
     PICa & Zero-shot & None & 54.3 \\
     Vicuna & 16-shots & Chain-of-Thought & 68.6 \\
     IMMO & 16-shots & Inner-Monologue & 74.0 \\
    \midrule
     Vicuna & SL & Chain-of-Thought & 78.3 \\
     IMMO & SL+RL & Inner-Monologue & \textbf{84.8} \\
    \bottomrule
  \end{tabular}
  }
  \caption{Results on ScienceQA.}
 \label{tab:result-sqa}
\end{table}

\subsection{Visual Entailment}

SNLI-VE~\citep{xie2018visual} is a widely-used VE task built on top of Stanford Natural Language Inference (SNLI)~\citep{bowman2015large} and Flicker30k~\citep{plummer2015flickr30k} image datasets. This task is designed as a classification problem for vision-language reasoning: identify whether the relationship between the given image premise and text hypothesis is entailment, neural, or contradiction.
% Solving SNLI-VE requires an understanding of detailed image information, as well as the ability to reason with commonsense. 

\subsubsection{Result}
Table~\ref{tab:result-ve} shows the results on the SNLI-VE dev set. We add another baseline: IdealGPT~\citep{idealgpt}, a recent hybrid integration approach that utilizes the rich reasoning knowledge of GPT-3.5-175B. Among approaches that use text to represent visual information, IMMO achieves the best performance. With a much smaller LLM, our best-performing checkpoints trained from Vicuna-7B achieved a 10.4\% improvement over IdealGPT (65.7\% vs 55.3\%). 

The results of 3 embedding-based methods (MiniGPT4 \citep{zhu2023minigpt}, LLaVA \citep{liu2023visual}, and OFA \citep{wang2022ofa}) are also reported as a reference. Well-tuned embedding-based methods such as OFA work extremely well on this dataset, illustrating the power of using a single model for end-to-end optimization. These results suggest that for tasks like SNLI-VE where natural language is not enough to describe necessary visual information, approaches that convert images into text for LLM reasoning are sub-optimal. However, in real-world practice, many companies choose not to perform single-model-based end-to-end (embedding) optimization because interpretability is required or the training cost is too high. In such situations, hybrid integration like IMMO could be a good choice.   

\begin{table}[t]
\centering
\resizebox{0.8\linewidth}{!}{
  % \centering
  \small
  \setlength{\tabcolsep}{3pt}
  \begin{tabular}{l l c }
    \toprule
     & Method & Accuracy(\%) \\
    \midrule
    \multirow{3}{*}{\rotatebox{90}{\color{gray}{Emb}}} & \color{gray}{MiniGPT4 \citep{zhu2023minigpt}} & \color{gray}{35.1} \\
    & \color{gray}{LLaVA (ZS) \citep{liu2023visual}} & \color{gray}{40.3} \\
    & \color{gray}{OFA \citep{wang2022ofa}} & \color{gray}{91.0} \\
    \midrule
    \multirow{4}{*}{\rotatebox{90}{NL}} & IdealGPT \citep{idealgpt} & 55.3 \\
    \cmidrule{2-3}
    & Vicuna-16-shots & 49.8 \\
    & Vicuna-SL & 59.8 \\
    & IMMO & 65.7 \\
    \bottomrule
  \end{tabular}
  }
  \caption{Results on SNLI-VE. The \textit{Emb} group includes embedding alignment approaches, while the \textit{NL} group includes methods using text to represent visual information.}
 \label{tab:result-ve}
\end{table}

%%%%%%%%%%%%%%%%%%%%%%%%%%%%%%%%%%%%%%%%%%%%%%%%%%%%%%%%%%%%%%
\section{Ablation Studies and Analysis}
\begin{figure*}[tbh]
    \centering
    \includegraphics[width=0.9\linewidth]{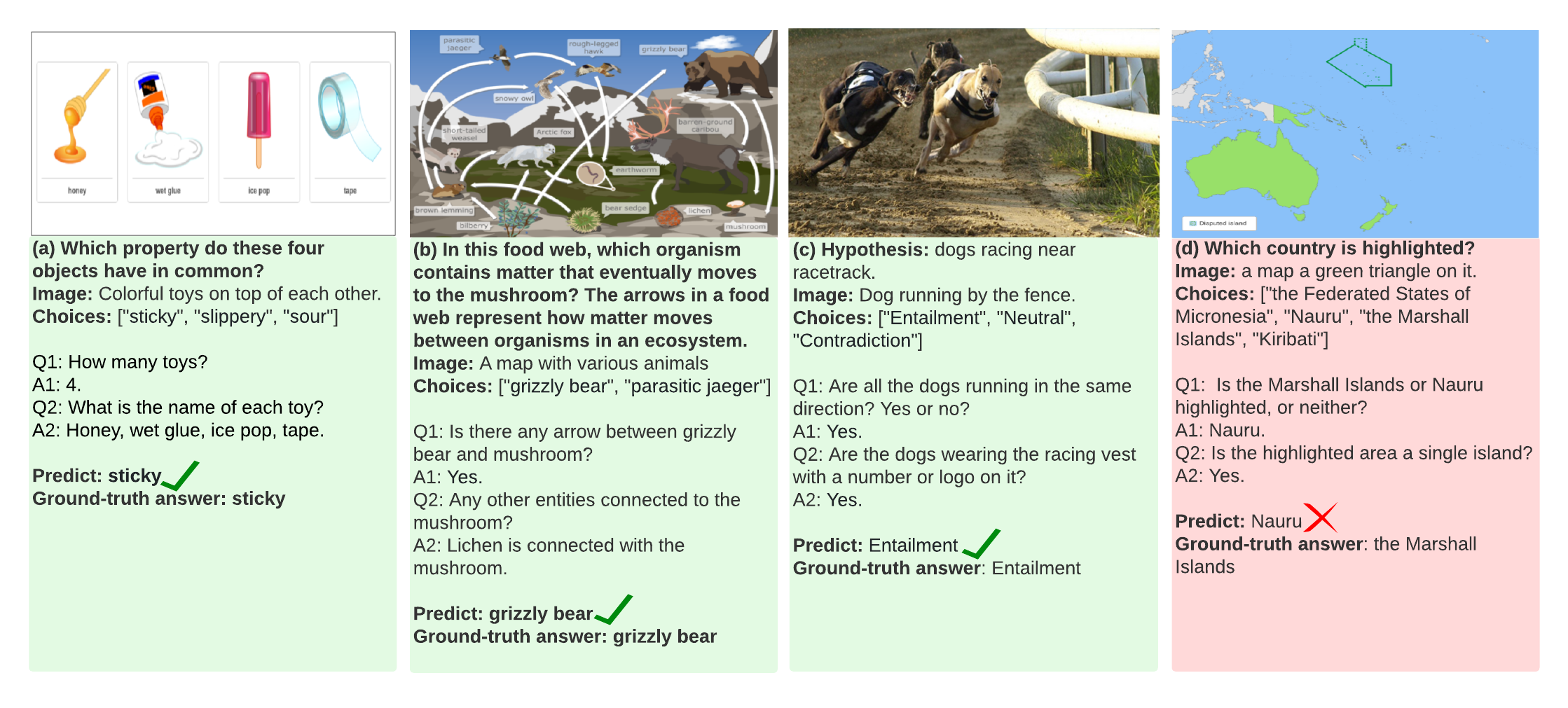}
    % \vspace{-2ex}
    \caption{Success and failure examples of IMMO.}\label{fig:sqa-example}
\end{figure*}

\subsection{Representative cases}
Figure~\ref{fig:sqa-example} displays instances of successful outcomes (a, b, and c) as well as an unsuccessful case (d) depicting the interpretable inner monologues generated by IMMO. Example (b) shows LLM's ability to compensate for VLM inaccuracies (incorrect A2) through reasoning and available information. Moreover, the questioning path in (b) and (c) demonstrate LLM's vigilance in monitoring VLM responses, persisting in using subsequent questions Q2 to validate information after Q1, even when the information is enough to answer the main question. 
On the other side, example (d) exposes how the VLM's limited geographical background knowledge hinders LLM from arriving at an accurate answer. However, the erroneous visual information from VLM misleads the LLM into an incorrect final prediction. These examples illustrate the interpretability of IMMO.

\subsection{Comparison with additional VQA methods}
We did some further analysis to compare representative solutions on the ScienceQA task (Table \ref{tab:result-sqa-with-SOTA}). The embedding alignment approach, LLaVA~\citep{liu2023visual}, performs well; however, it requires extensive training data and lacks interpretability. Hybrid integration such as IMMO, Chamaleon ~\citep{lu2023chameleon} and UnifiedQA~\citep{khashabi2020unifiedqa, lu2022learn} employ modular architecture, enabling model-wise interpretability by access to individual outputs from sub-modules. Chameleon is based on GPT-4, which is not publicly available, and poses constraints on its adoption and further fine-tuning. Compared with Chamaleon, IMMO achieved comparable performance with a significantly smaller model. UnifiedQA uses supervised training akin to our Vicuna-SL baseline, however, it falls short due to the lack of system-wide optimization and information loss when converting images to captions. Compared with UnifiedQA, IMMO addresses these problems via inner monologue and two-stage training, which significantly improves the performance of the hybrid integration method. Furthermore, compared to Chamaleon and UnifiedQA, which offers simple model-level interpretability, IMMO's entire complex multi-round reasoning procedure is also transparent and human-readable. %One can inspect the inner monologue for bad spots in the reasoning chain to identify the causes of undesirable results.

Notably, we can not rule out the possibility that black-box models like GPT-4 might have inadvertently or intentionally undergone training using publicly accessible test data. Thus we list the performance of those black-box models here for reference instead of as a baseline for a fair comparison. We expect the proposed approach could be applied to other LLMs and VLMs such as GPT-4 and further improve their performance. 

% As shown in the results in table~\ref{tab:result-sqa}, under the same training data usage (few-shot), the system gains significant improvement by allowing the language model to interact to gather desired visual information. Our few-shot conversational approach also outperforms the supervised fine-tuned baseline on the SQA. Even using all the training data, achieving good results solely by the language model is challenging due to the lack of ideal image information. Compared to this, we observe a decent improvement using the system-level RL optimization.
% Compared to other hybrid integration works, we are the first work that applies system-level optimization. Our proposed approach achieves comparable results compared to the GPT-4 based approach with a relatively smaller model. Visual alignment methods such as LLaVa have achieved very impressive results, however, learning alignment requires a small amount of training data. Also, interpretation of the results is very difficult. Note that MM-CoT heavily relies on the human-written Chian-of-Thought, which we can see as the pre-written inner monologue. However, in most real-world visual language use cases, it can hardly find such high-quality manually-written CoTs, thus limiting the chances of the model being widely used. Our system, on the other hand, with the interactively auto-generate of the inner monologue does not require any mid-step annotation on the task. With RL optimization, a decent performance can be achieved using only pair question-answer data. 

\begin{table}[tb]
\resizebox{\linewidth}{!}{
  \centering
  \setlength{\tabcolsep}{3pt}
  \begin{tabular}{|l|c|c|c|c|}
    \hline
     & LLaVa & Chamaleon & UnifiedQA & IMMO \\
    \hline
    Interpretable & \color{red}{\ding{55}} & \color{teal}{\ding{51}} & \color{teal}{\ding{51}} & \color{teal}{\ding{51}} \\
    Trainable & \color{teal}{\ding{51}} & \color{red}{\ding{55}} & \color{teal}{\ding{51}} & \color{teal}{\ding{51}} \\
    Model size & 13B & GPT-4 & 223M & 9B \\
    Tuned param & 13B & 0 & 223M & 5M \\
    Data usage & 770K & - & 17K & 25K \\
    SQA & 90.9 & 86.5 & 74.1 & 84.8 \\
    \hline
  \end{tabular}
  }
  \caption{Comparison with other ScienceQA approaches.}
 \label{tab:result-sqa-with-SOTA}
\end{table}

% \begin{table*}[t]
%  \resizebox{\linewidth}{!}{
%  \centering
%  \setlength{\tabcolsep}{3pt}
%  \begin{tabular}{l l l l l l}
%    \toprule
%    Method & Data & Model size & Tuned & Interpretable & SQA \\
%    \midrule
%    %LLaVA (GPT-4) & 753K+17K & - & 13B & No & 92.5 \\ % 595K+158K
%    LLaVA & 753K+17K & 13B & 13B & No & 90.9 \\
%    MM-CoT & 17K & 738M & 738M & No & 91.7 \\
%    LLaMA-adapter &  &  &  & No & 85.2 \\
%    VisualBert  &  &  &  & No & 61.9 \\
%    \midrule
%    IMMO & 17K+8K & 9B & 5M & M+R & 84.8 \\
%    \bottomrule
%  \end{tabular}
%  }
%  \caption{Embedding alignment works on ScienceQA}
% \label{tab:result-sqa-with-SOTA}
%\end{table*}

\subsection{The Impact of Conversation Turns}
\begin{figure}[tb]
    \centering
    \includegraphics[width=0.9\linewidth]{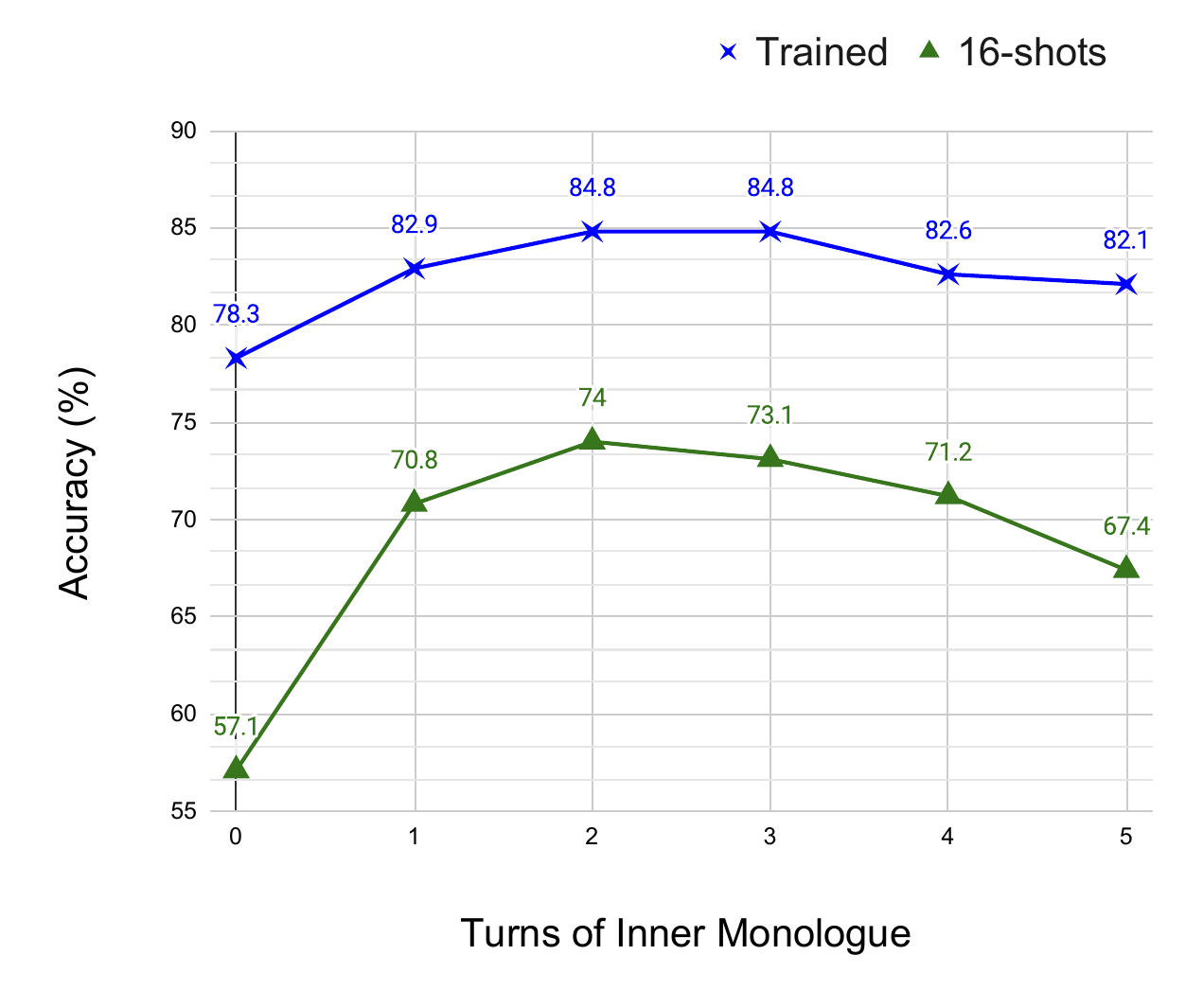}
    % \vspace{-2ex}
    \caption{Ablation study on different inner monologue turns using on ScienceQA test set under few-shot and trained manner.
    }\label{fig:ablation_turns}
\end{figure}

To examine the impact of inner monologue turns on performance, we conduct ablation tests on ScienceQA using both few-shot and trained approaches. Maintaining constant hyperparameters, we evaluate turns ranging from 0 to 5, where 0 means VLM only provides an initial caption. As shown in Figure~\ref{fig:ablation_turns}, accuracy notably rises on the SQA test set from 0 to 2 turns, plateauing thereafter. Our analysis identifies SQA questions as demanding less multi-hop reasoning than background knowledge. Thus, LLM's primary learned strategy involves querying key facts initially, followed by confirmation or asking for side information. Also, more conversations bring uncertainty to the interaction as LLM may ask less relevant questions after 3 turns or VLM brings incorrect visual information.
This trend is accentuated under the few-shot setting. Without training, LLM appears to be less robust to noise conversation, so the performance rapidly decreases after 2 turns. It's important to note that these findings are specific to ScienceQA question patterns, underscoring the best inner monologue turns are highly based on the dataset's characteristics.

%%%%%%%%%%%%%%%%%%%%%%%%%%%%%%%%%%%%%%%%%%%%%%%%%%%%%%%%%%%%%%
\section{Conclusion and Future Work}
Inspired by cognitive modeling, we apply inner monologue, a commonly seen human reasoning process, in the interaction between LLM and VLM. We proposed to learn which questions to ask and how to answer questions during the multi-round monologue using a two-stage training framework together with a newly constructed training corpus from existing VQA datasets. Our experiments demonstrated the ability to learn how to do inner monologues, as well as the effectiveness of acquiring information and reasoning through inner monologues.

This paper is a first step towards this research direction, and there is much room for future improvement. Our current implementation promotes the reasoner querying certain turns, while an ideal reasoner should autonomously determine whether to continue querying or end the inner monologue with direct answers (for example, when adequate information has been gathered or due to time/resource constraints). Our implementation only includes one observer, while it's possible to include more observers with different modalities or functionalities. Due to the resource limits, we used a synthetic way to generate supervised training data, while organizations with ample resources could hire human annotators could provide more labeled data with higher quality. The reward function could also be further studied.
%Our implementation used exact matching between the predicted answer and ground truth answer as reward, while other reward function such as semantic match or model based reward function could be used in the future.
More importantly, our proposed approach is an automated way of generating intermediate steps Chain-of-Thought, and we expect the concept of inner monologue can be widely applied to a variety of use cases. %We tried some simple prompt to let LLM decide, however the results are not successful due to the well-known challenges of reinforcement learning within natural language processing, where the sheer vastness of the language space presents a formidable hurdle in the quest for optimal policy. From our experiments, we observe that granting the language model the discretion to choose between asking and answering hampers the convergence of the RL training, primarily due to the exponential increase in the size of the action space. To address this dilemma, one potential avenue is to use a stronger LLM as starting point. More promisingly, we expect better RL algorithms or more sophisticated reward function designs to provide promising avenues for alleviating this limitation.

%%%%%%%%%%%%%%%%%%%%%%%%%%%%%%%%%%%%%%%%%%%%%%%%%%%%%%%%%%%%%%

% \appendix

\bibliography{aaai24}

\end{document}